\documentclass[letterpaper,10pt,conference]{ieeeconf}
\IEEEoverridecommandlockouts
\usepackage{amsmath,amssymb}
\usepackage{graphicx}
\usepackage{booktabs}
\usepackage{cite}
\usepackage{url}
\usepackage[nocheckfootnote]{flushend}
\usepackage{needspace}
\usepackage{etoolbox}
  \usepackage[T1]{fontenc}
  \usepackage{mathptmx}
  \patchcmd{\abstract}{\bfseries\textit{Abstract}}{\normalfont Abstract}{}{}

\usepackage[hidelinks]{hyperref}
\hypersetup{pdftitle={DCLP++: Learning to Navigate with Footprint Clearance and Relative Motion}}
\newcommand{\method}{DCLP++}
\newcommand{\clip}{\operatorname{clip}}
\newcommand{\dist}{\operatorname{dist}}

\newcommand{\R}{\mathbb{R}}
\newcommand{\B}{\mathcal{B}}

\newcommand{\p}{\mathbf{p}}
\newcommand{\q}{\mathbf{q}}
\newcommand{\uvec}{\mathbf{u}}
\newcommand{\z}{\mathbf{z}}
\newcommand{\fvec}{\mathbf{f}}
\newcommand{\svec}{\mathbf{s}}
\newcommand{\vvec}{\mathbf{v}}
  \author{Shanze Wang$^{1,2}$ and Wei Zhang$^{1,\dagger}$%
\thanks{This work was supported in part by the National Natural Science Foundation of China under Grant 62503251.}%
\thanks{$^{1}$Shanze Wang and Wei Zhang are with the College of Information Science and Technology, Eastern Institute of Technology, Ningbo, China.
{\tt\footnotesize szwang@eitech.edu.cn, zhw@eitech.edu.cn}}%
\thanks{$^{2}$Shanze Wang is also with the Department of Aeronautical and Aviation Engineering, The Hong Kong Polytechnic University, Hung Hom, Kowloon, Hong Kong.
{\tt\footnotesize shanze.wang@connect.polyu.hk}}%
\thanks{$^{\dagger}$Corresponding author: Wei Zhang.}%
}

  \hypersetup{pdfauthor={Shanze Wang and Wei Zhang}}

\title{\LARGE DCLP++: Learning to Navigate with\\Footprint Clearance and Relative Motion}

\begin{document}
\maketitle
\thispagestyle{empty}
\pagestyle{empty}

\begin{abstract}
We present \method{}, a local navigation framework that uses footprint clearance as the geometric basis for studying relative motion features in dynamic environments. Each valid LiDAR return is mapped to its shortest Euclidean distance from the filled robot footprint before reciprocal encoding, replacing distance from the sensor with distance to the occupied body. Radial measurements or simulated planar relative velocities provide short-horizon features without static--dynamic labels in the policy input. A preliminary study uses a rectangular robot with a speed limit of $1\,\mathrm{m/s}$ among 20 moving obstacles. On 100 fixed validation tasks, two selected training seeds yield mean success rates of 42\% with sensor range and 70\% with footprint clearance after 200{,}000 environment steps. Motion variants show mixed additional gains. These results support the clearance-based observation in the evaluated setting; reliable motion benefits and transfer across robots require further evaluation.
\end{abstract}

\section{Introduction}
Safe and efficient navigation in environments shared with people and other moving agents requires mobile robots to avoid collisions while making progress toward their goals~\cite{drlvo,dsrnn}. Static structures restrict the available space, while obstacle motion and interactions require decisions from incomplete observations~\cite{drlvo,dsrnn,sicnav}. Deep reinforcement learning (DRL) has enabled navigation policies to respond to such conditions in both ground navigation and autonomous flight~\cite{drlvo,navrl}. These advances motivate the study of observations that relate the surrounding geometry and motion to the space occupied by the robot, so that the policy can reason about both current separation and approaching obstacles.

Local navigation methods address this problem through different forms of geometric and motion reasoning. Velocity-space methods use obstacle positions and velocities to identify collision-producing actions, with reciprocal formulations also accounting for mutual avoidance~\cite{vo,orca}. Predictive optimization can couple the planned robot motion with models of how nearby agents respond~\cite{sicnav}. Learning-based methods instead allow a policy to map sensor measurements to actions through experience~\cite{long}. Although their decision mechanisms differ, these approaches all depend on how the observation describes the relationship between the robot and nearby obstacles. For a learned policy, this raises a practical design question about which geometric and motion quantities should be supplied before feature extraction.

A single range scan does not distinguish obstacles that approach, recede, or cross the intended path from the same observed positions. Existing navigation systems address this ambiguity by providing estimated object states~\cite{drlvo,navrl}, extracting information from scan sequences~\cite{tagd,lndnl}, or combining distances with point flow~\cite{p2m}. Motion information can therefore be obtained at the object level or from changes in the observed surfaces. The latter supports navigation without requiring every obstacle to be represented as a tracked object~\cite{tagd,p2m}. This motivates expressing motion at the return-point level, where a static surface and a moving object can both be described by their velocity relative to the robot.

The geometric reference of these motion features also matters. LiDAR range measures distance from the sensor, whereas collision depends on the region occupied by the robot. For a rectangular body, equal ranges in different directions can leave different gaps to its boundary. Reciprocal distance encoding can emphasize near-obstacle differences~\cite{ipaprec}, while dimension-conditioned inputs and observation--action scaling address changes in robot configuration~\cite{dclp,dvst}. Point-to-robot distances have also been used in model-based collision constraints~\cite{neupan}. These findings suggest computing the known relationship between a return point and the footprint before learning. Relative motion can then be evaluated against the same occupied region that defines current separation.

Building on dimension-configurable local navigation~\cite{dclp}, we present \method{}, a framework that uses footprint clearance as the geometric basis for studying motion-aware observations. Each return is expressed in the robot frame and assigned its shortest Euclidean distance to the filled footprint. A reciprocal transform encodes this clearance, and short-horizon point rollouts provide motion features relative to the same footprint. The policy interface requires no static--dynamic class labels. A clearance-only variant isolates the change from sensor range to body clearance; radial and simulated planar velocity variants examine the additional role of motion information.

This paper provides a reproducible clearance construction, specifies motion features on that geometric basis, and evaluates their respective effects through observation comparisons. The preliminary study uses a rectangular robot with a speed limit of $1\,\mathrm{m/s}$ among 20 moving obstacles. On fixed validation tasks, clearance improves success over sensor range in two selected training runs, while the additional gains from motion features vary across formulations and runs. The resulting geometric interface provides a basis for subsequent studies of dynamic navigation across robot configurations, with the present evaluation focused on the fixed-footprint setting.

\begin{figure*}[t]
\centering
\includegraphics[width=\textwidth]{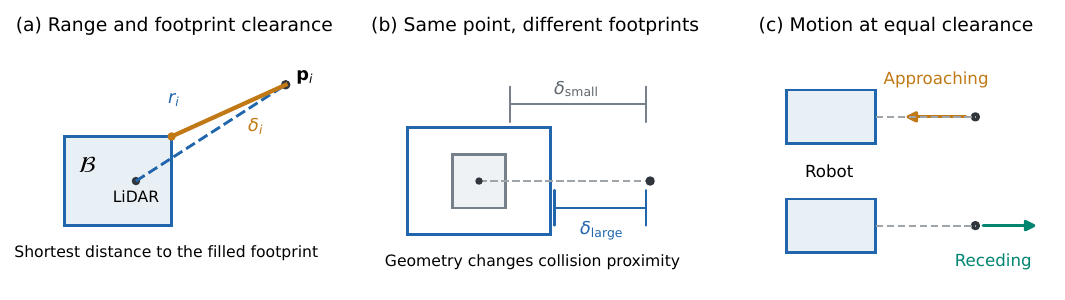}
\caption{Observation geometry and motion. (a) Sensor range $r_i$ and footprint clearance $\delta_i$ can differ in length and direction. The illustrated sensor offset explains the general transform; experiments use a centered sensor. (b) Clearance depends on the footprint. (c) Equal clearance can precede different motion. These are definition sketches, not experimental trajectories.}
\label{fig:geometry}
\end{figure*}

\section{Related Work}
\subsection{Model-Based Local Planning in Dynamic Environments}
Model-based local planners express obstacle avoidance through geometric and motion constraints. Dynamic-window search restricts candidate velocities according to acceleration bounds and stopping feasibility~\cite{dwa}. Velocity obstacles describe collision-producing velocities under an assumed obstacle motion~\cite{vo}, while reciprocal formulations account for avoidance by other agents~\cite{rvo,orca}. More recent model predictive control jointly optimizes robot motion and crowd predictions, using an explicit interaction model within a bilevel problem~\cite{sicnav}. These methods make current separation, relative velocity, and motion feasibility part of the planning formulation. Our work draws on these quantities to construct policy observations and studies their effects within a DRL local planner.

\subsection{Learning-Based Navigation in Dynamic Environments}
Learning-based navigation has developed along both sensor-level and agent-level routes. Sensor-level policies map range measurements to velocity commands without requiring the states of all surrounding agents~\cite{long}. Agent-level methods use recurrent processing for variable numbers of neighbors~\cite{everett}, attention to model crowd interactions~\cite{sarl}, and spatial--temporal graph structures to reason under partial observability~\cite{dsrnn}. Other systems combine scan history with pedestrian kinematics~\cite{drlvo}, or process static occupancy and tracked dynamic-object states through separate feature extractors~\cite{navrl}. These choices determine whether obstacle motion is supplied by a perception module or must be inferred from the observations received by the policy.

Temporal sensor representations provide another way to describe motion. Aligned scan groups with spatial--temporal attention support anticipatory navigation without explicit object tracking~\cite{tagd}, and recurrent processing of normalized LiDAR scans provides information across successive observations~\cite{lndnl}. Stochastic occupancy prediction represents possible future scenes while accounting for robot motion, moving objects, and static geometry~\cite{scope}. Point flow has also been combined with a distance map to support dynamic flight from LiDAR observations~\cite{p2m}. These studies already establish alternatives to a separate object-tracking pipeline. Our focus is the geometric reference used with motion information: current and projected point distances are measured to the occupied robot footprint, and their effect on policy learning is evaluated separately from sensor range.

\subsection{Robot Geometry and Transfer in Learned Navigation}
Observation design also affects how a policy handles changes in robot geometry. Learnable reciprocal transforms allocate greater numerical variation to nearby obstacles~\cite{ipaprec}. Dimension-conditioned point features and a configuration-dependent training curriculum allow a local policy to handle rectangular robots of different sizes and motion limits~\cite{dclp}. Observation and action scaling offers another route to dimension-variable navigation~\cite{dvst}. Across broader robot classes, policies have been trained through expert distillation over varied embodiments~\cite{xnav}, while geometric action generation has been separated from learned adaptation to target dynamics~\cite{cenav}. These approaches address different aspects of transfer, from the geometry of the observation to the motion produced by a control command.

Robot geometry is also represented explicitly in optimization. Point-level distance constraints connect raw obstacle observations to robot shape~\cite{neupan}; multi-frame extensions add predicted points for dynamic navigation~\cite{mfneupan}. Swept-volume signed distance fields further account for the geometry occupied over a continuous trajectory~\cite{svsdf}. We use the established point-to-footprint distance as a feature for a learned policy and evaluate short-horizon motion features defined on the same region. Recomputing this distance for another footprint gives a common geometric definition, while policy transfer additionally requires suitable training coverage and control constraints. The experiments here isolate the effect of the clearance input for one rectangular footprint.

\section{Footprint Clearance and Relative Motion Representation}
\label{sec:method}
\subsection{Observation and Robot Geometry}
For planar goal-directed navigation, let $\B\subset\R^2$ be the closed, filled robot footprint in the body frame. A return with range $r_i$ and sensor-frame bearing $\theta_i$ is transformed using the known sensor-to-body rotation $R_{BL}$ and translation $\mathbf t_{BL}$:
\begin{equation}
\p_i=R_{BL}r_i[\cos\theta_i,\sin\theta_i]^\top+\mathbf t_{BL}.
\label{eq:point}
\end{equation}
The body-frame bearing is $\vartheta_i=\operatorname{atan2}(p_{i,y},p_{i,x})$. Experiments use a centered, aligned sensor, so $\vartheta_i=\theta_i$; other mounting positions are covered by~\eqref{eq:point} but are not evaluated.

Goal distance and bearing $(d_g,\varphi_g)$, current velocity $(v,\omega)$, and linear and angular speed and acceleration limits form the additional state
\begin{equation}
\svec=[d_g,\varphi_g,v,\omega,v_{\max},\omega_{\max},a_{v,\max},a_{\omega,\max}].
\label{eq:state}
\end{equation}
The policy maps point features and $\svec$ to bounded velocity commands. Points describe visible surfaces; occluded geometry remains unobserved.

\subsection{Footprint Clearance Encoding}
\label{sec:clearance}
For each valid return, we define
\begin{equation}
\delta_i=\dist(\p_i,\B)=\min_{\q\in\B}\|\p_i-\q\|_2.
\label{eq:clearance}
\end{equation}
This unsigned distance is positive outside the footprint and zero on its boundary or inside it. For a centered rectangle $\B=[-a,a]\times[-b,b]$,
\begin{equation}
\delta_i=\sqrt{\max(|p_{i,x}|-a,0)^2+\max(|p_{i,y}|-b,0)^2}.
\label{eq:rectangle}
\end{equation}
The experimental robot has $a=0.30\,\mathrm m$ and $b=0.25\,\mathrm m$. A centered circle of radius $R$ gives $\max(\|\p_i\|_2-R,0)$. For polygons, exterior points use the minimum distance to boundary segments; interior points receive zero.

For rectangles, subtracting a constant radius from range is generally incorrect: the closest point can lie on a side or corner, away from the measurement ray (Fig.~\ref{fig:geometry}). Clearance supplies this geometry before learning, but distances to visible samples do not certify collision freedom for unobserved surfaces or future motion.

The baseline clearance variant, B0, replaces range in an IPAPRec-style point feature while preserving the reciprocal map:
\begin{equation}
\z_i=[\cos\vartheta_i,\sin\vartheta_i,\phi_\alpha(\delta_i)],
\qquad
\phi_\alpha(d)=\frac{1}{\max(d+\alpha,\epsilon)}.
\label{eq:encoding}
\end{equation}
Distances are in meters. Actor and critic learn separate scalar offsets $\alpha$, initialized to zero. The lower clip at $\epsilon=10^{-8}$ uses a straight-through derivative. This is the reciprocal form in~\cite{ipaprec} with the offset sign reversed. B0 excludes obstacle-motion features but retains robot velocity in~\eqref{eq:state}.

All variants divide 540 beams into 90 contiguous angular bins, select the valid beam with minimum original range in each bin, and retain its direction and clearance. An empty bin uses its mean bearing and a virtual maximum-range point as a finite fallback, without future-motion evidence. This fallback does not indicate a detected obstacle.

A downstream interface can instead compute clearance before sector-wise minimum pooling and assign a maximum-distance sentinel to empty sectors. Range-based and clearance-based selection generally differ. Our results use the former; implementations reusing the definition must specify their pooling rule, missing-return convention, and numerical floor.

\subsection{Relative Motion Features}
\label{sec:motion}
For a physical obstacle point with world-frame velocity $\vvec_i^W$, let $\uvec_i=R_{WB}^\top(\vvec_i^W-\vvec_r^W)$ denote relative translational velocity expressed in the current body axes. Its body-frame coordinate derivative satisfies
\begin{equation}
\dot{\p}_i=\uvec_i-\omega J\p_i,
\qquad J=\begin{bmatrix}0&-1\\1&0\end{bmatrix}.
\label{eq:relative}
\end{equation}
Static points have $\vvec_i^W=0$ but can have nonzero relative velocity. Class labels are unnecessary in this interface; reliable motion acquisition remains a sensing problem.

With the centered sensor, F1 uses the line-of-sight measurement $\dot r_i$ to form $\tilde\uvec_i=\dot r_i[\cos\theta_i,\sin\theta_i]^\top$; $\dot r_i<0$ indicates approach. This radial component omits tangential motion and generally differs from the clearance derivative. G2 receives planar relative velocity from simulator hit metadata as an oracle diagnostic input.

Both motion variants use a fixed-orientation linear rollout,
\begin{equation}
\p_i(\tau_k)=\p_i+\hat\uvec_i\tau_k,
\qquad
\delta_i(\tau_k)=\dist(\p_i(\tau_k),\B),
\label{eq:rollout}
\end{equation}
where $\hat\uvec_i$ is the radial approximation or oracle velocity and $\tau_k=k\Delta\tau$, $k=1,\ldots,K$. This feature approximation omits the rotation term in~\eqref{eq:relative}, future control changes, and obstacle acceleration.

F1 uses $\Delta\tau=0.1\,\mathrm s$ and $K=8$. Let $\delta_{i,\min}$ be the minimum sampled clearance and $\tau_i^*$ its earliest occurrence. The motion gate and modified distance feature are
\begin{equation}
g_i=\frac{\beta_g}{\delta_{i,\min}+\beta_g}
\exp\!\left(-\frac{\tau_i^*}{T_g}\right),
\quad
q_i=\phi_\alpha(\delta_i)(1+\lambda g_i),
\label{eq:gate}
\end{equation}
with $\beta_g=0.1\,\mathrm m$ and $T_g=0.8\,\mathrm s$. The factor $\lambda$ increases linearly from zero to one over 30{,}000 soft actor-critic (SAC) updates. Direction and $q_i$ form the encoder input; no separate velocity channel is added.

G2 uses $K=4$ and a $0.4\,\mathrm s$ horizon. It computes $c_i$, an indicator of zero clearance at any sampled future time, and $t_{e,i}$, the first such time. When no entry occurs, $c_i=0$ and $t_{e,i}=0.4\,\mathrm s$. Its future feature is
\begin{equation}
\begin{split}
\fvec_i&=[c_i,1-\bar\delta_{i,\min},c_i(1-\bar t_{e,i})],\\
\bar\delta_{i,\min}&=\clip(\delta_{i,\min}/(1\,\mathrm m),0,1),\\
\bar t_{e,i}&=\clip(t_{e,i}/(0.4\,\mathrm s),0,1).
\end{split}
\label{eq:triplet}
\end{equation}
A linear point layer maps $\fvec_i$ to 32 channels, applies the same warmup factor, and adds the result to the first current-clearance layer. Entry between discrete samples can be missed.

\begin{figure*}[t]
\centering
\includegraphics[width=\textwidth]{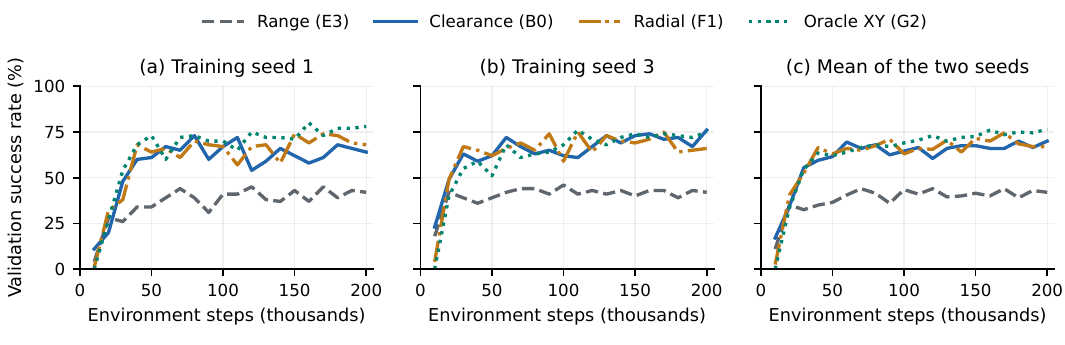}
\caption{Offline success on the same 100 tasks at 20 checkpoints for two selected seeds and their mean. Curves connect measurements without smoothing. Oracle XY uses simulator-provided planar relative velocity. Results do not estimate performance across all attempted seeds.}
\label{fig:curves}
\end{figure*}

\subsection{Policy and Training}
Shared $1\times1$ convolutions with 32, 64, and 128 channels and max pooling over points form a PointNet-style encoder~\cite{pointnet}. Its output is concatenated with~\eqref{eq:state} and processed by four fully connected layers of width 128. The actor outputs a four-component Gaussian mixture over two-dimensional normalized actions. SAC~\cite{sac} uses two action-value estimates and separate actor and critic encoders.

For $(a_v,a_\omega)\in[-1,1]^2$, target velocities are $v_{\mathrm{target}}=(a_v+1)v_{\max}/2$ and $\omega_{\mathrm{target}}=a_\omega\omega_{\max}$, with changes bounded by acceleration limits and the control interval. Nonterminal reward is twice the decrease in goal distance; terminal rewards are $+10$ for arrival and $-10$ for collision or leaving the world. Arrival requires goal distance below $0.2\,\mathrm m$ without collision. All variants share these settings.

\section{Preliminary Evaluation}
\subsection{Experimental Protocol}
IR-SIM experiments use a $20\times20\,\mathrm m$ world and a fixed $0.6\times0.5\,\mathrm m$ differential-drive robot, with velocities in $[0,1]\,\mathrm{m/s}$ and $[-\pi,\pi]\,\mathrm{rad/s}$. Per-episode acceleration limits are uniform in $[0.5,5]\,\mathrm{m/s^2}$ and $[\pi/6,2\pi]\,\mathrm{rad/s^2}$. Simulation, control, and scanning run at 10~Hz for at most 400 steps. Interior starts and goals have at least $2\,\mathrm m$ separation and $0.5\,\mathrm m$ initial obstacle clearance.

Twenty circular or polygonal obstacles use reciprocal velocity obstacle (RVO) control~\cite{rvo}: twelve consider the robot, while eight ignore it but avoid other obstacles. Each task selects a preferred-speed limit from $\{1,2,3\}\,\mathrm{m/s}$ and samples obstacle speeds uniformly from zero to that limit; RVO adjusts actual velocities. The 100 validation tasks allocate 33, 33, and 34 cases to these limits. Obstacle geometry is kept inside the world.

Simulated frequency-modulated continuous-wave (FMCW) LiDAR provides $360^\circ$ coverage, 540 beams, and a $0.02$--$15\,\mathrm m$ range. Noise combines a uniform term in $[-0.01,0.01]\,\mathrm m$ and a zero-mean Gaussian term with standard deviation of 1\% of clean range. Radial-velocity noise and scan motion compensation are disabled; physical sensing is not validated.

Training lasts 200{,}000 environment steps, with checkpoints every 10{,}000 steps, discount 0.99, actor and critic learning rates $10^{-4}$, batch size 100, entropy coefficient 0.01, and replay capacity of two million transitions. A PID controller supplies the first 101 episodes. Offline evaluation reuses task seeds 4{,}000{,}001--4{,}000{,}100 and executes the squashed mean of the highest-logit Gaussian component without sampling.

The archive contains seeds 1 and 3, evaluated on CUDA and CPU, respectively. They were selected because other runs exhibited training collapse. Results therefore describe selected runs, not stability across all attempted seeds; checkpoints evaluated on reused tasks are not independent training samples.

\subsection{Effect of Footprint Clearance}
E3 uses direction and reciprocal sensor range without obstacle motion. B0 changes only the distance geometry, retaining beam selection, encoder, reciprocal map, and training settings. At the final checkpoint (Table~\ref{tab:results}), B0 reaches 64\% and 76\%, versus 42\% for both E3 seeds, a mean gain of 28 percentage points. The curves remain separated through much of training (Fig.~\ref{fig:curves}).

\begin{table}[t]
\caption{Success rate (\%) at 200{,}000 environment steps on the fixed validation set. Means cover only the two selected seeds.}
\label{tab:results}
\centering
\small
\begin{tabular}{lrrr}
\toprule
Variant & Seed 1 & Seed 3 & Mean \\
\midrule
Range (E3) & 42.0 & 42.0 & 42.0 \\
Clearance (B0) & 64.0 & 76.0 & 70.0 \\
Power reciprocal (B0-Prec) & 42.0 & 36.0 & 39.0 \\
Radial motion (F1) & 68.0 & 66.0 & 67.0 \\
Oracle motion (G2) & 78.0 & 75.0 & 76.5 \\
\bottomrule
\end{tabular}

\end{table}

The comparison supports supplying body clearance directly under this training protocol. The fixed-footprint evaluation does not test policy transfer.

B0-Prec retains the clearance input and changes the reciprocal parameterization to
\begin{equation}
\phi_{\beta,p}(d)=\frac{(\beta+0.5)^{p-1}}{(d+\beta)^p},
\qquad \beta>0.1,\quad 1<p<2,
\label{eq:power}
\end{equation}
where $d$ and $\beta$ are numerical distances in meters. With initial $\beta=0.5$ and $p=1.1$, mean success is 39\%; the B0 gain thus depends on the encoding as well as the distance geometry.

\subsection{Effect of Motion Information}
F1 reaches 67\% mean final success, three percentage points below B0. G2 reaches 76.5\%, improving over B0 by 14 points in seed 1 but decreasing by one point in seed 3. Its feature-fusion layer and shorter horizon also differ from F1, so the comparison does not isolate velocity dimensionality.

Clearance gives the clearest improvement in these runs; motion benefits remain formulation-dependent. Tangential motion loss, fixed-orientation rollouts, feature scaling, and training variability are hypotheses for further testing, not established explanations of the differences.

\Needspace{5\baselineskip}
\section{Discussion and Conclusion}
\method{} provides a reproducible footprint-clearance observation and studies relative motion on the same geometric basis. Clearance improves validation success in the selected runs, while motion features yield mixed additional gains. Further evaluation must cover all attempted training seeds, physical motion sensing, and varied robot configurations. Turning motion and sensing uncertainty also require study. Neither achieved mean speed nor sensing-to-action latency is measured here; the configured speed limit and control rate do not establish these quantities. This evaluation scope defines the remaining work toward reliable dynamic navigation and transfer across robots.

\bibliographystyle{ieeetr}
\bibliography{references}
\end{document}